\documentclass[letterpaper, 10 pt, conference]{ieeeconf}  

\IEEEoverridecommandlockouts                              

\usepackage{times}
\usepackage{amsfonts} 
\usepackage{graphicx}
\usepackage{tabularx}
\usepackage{comment}
\usepackage{wrapfig}
\usepackage{amsmath}
\usepackage{amssymb}
\usepackage{array}
\usepackage{wrapfig,lipsum,booktabs}
\usepackage{hyperref}
\usepackage{booktabs}
\usepackage{multirow}
\usepackage{subfig}
\usepackage{setspace}
\usepackage{xcolor}

\usepackage{algorithm}
\usepackage{algorithmic}

\usepackage{hyperref}

\makeatletter
    \let\NAT@parse\undefined
\makeatother
\usepackage[square,numbers,sort&compress]{natbib}
\setcitestyle{numbers}
\usepackage{multicol}

\usepackage{titlesec}

\usepackage{titlesec}

\newcommand{\revision}[1]{{\color{black} #1}}

\begin{document}


\title{\LARGE \bf
DINOBot: Robot Manipulation via Retrieval and Alignment\\with Vision Foundation Models
}


\author{Norman Di Palo and Edward Johns \thanks{The Robot Learning Lab at Imperial College London. Email to: \texttt{n.di-palo20@imperial.ac.uk}}}



%

\maketitle

\begin{abstract}
   We propose DINOBot, a novel imitation learning framework for robot manipulation, which leverages the image-level and pixel-level capabilities of features extracted from Vision Transformers trained with DINO. When interacting with a novel object, DINOBot first uses these features to retrieve the most visually similar object experienced during human demonstrations, and then uses this object to align its end-effector with the novel object to enable effective interaction. Through a series of real-world experiments on everyday tasks, we show that exploiting both the image-level and pixel-level properties of vision foundation models enables unprecedented learning efficiency and generalisation. Videos and code are available at \href{https://www.robot-learning.uk/dinobot}{https://www.robot-learning.uk/dinobot}.
\end{abstract}


\section{Introduction}

The recent major successes in Deep Learning all had two main common ingredients: enormous, web-scale datasets, and substantial computational power to train increasingly large neural networks. However no such dataset of comparable size is available for robotics, where sensory perception must be coupled with actions. 

To take advantage of these large, pre-trained networks, the robotics community has often used them as backbone representations, to then train a neural network to predict actions on top of the extracted representations. However, imitation learning (IL) using these foundation models often still requires a considerable number of demonstrations for generalisation to emerge \cite{nair2022r3m, shah2021rrl, radosavovic2023real, seo2023masked}.

We argue that, rather than integrating foundation models into existing imitation learning methods as a backbone representation, we can design new imitation learning frameworks around these new capabilities of foundation models. To this end, we introduce \textbf{DINOBot}, a new imitation learning framework for robot manipulation tasks, which leverages the key capabilities of Vision Transformers (ViTs) trained through DINO \cite{caron2021emerging} (which we call DINO-ViTs), a self-supervised method for training vision networks.

\begin{figure}[t]
    \begin{center}
    \includegraphics[width=0.4\textwidth]{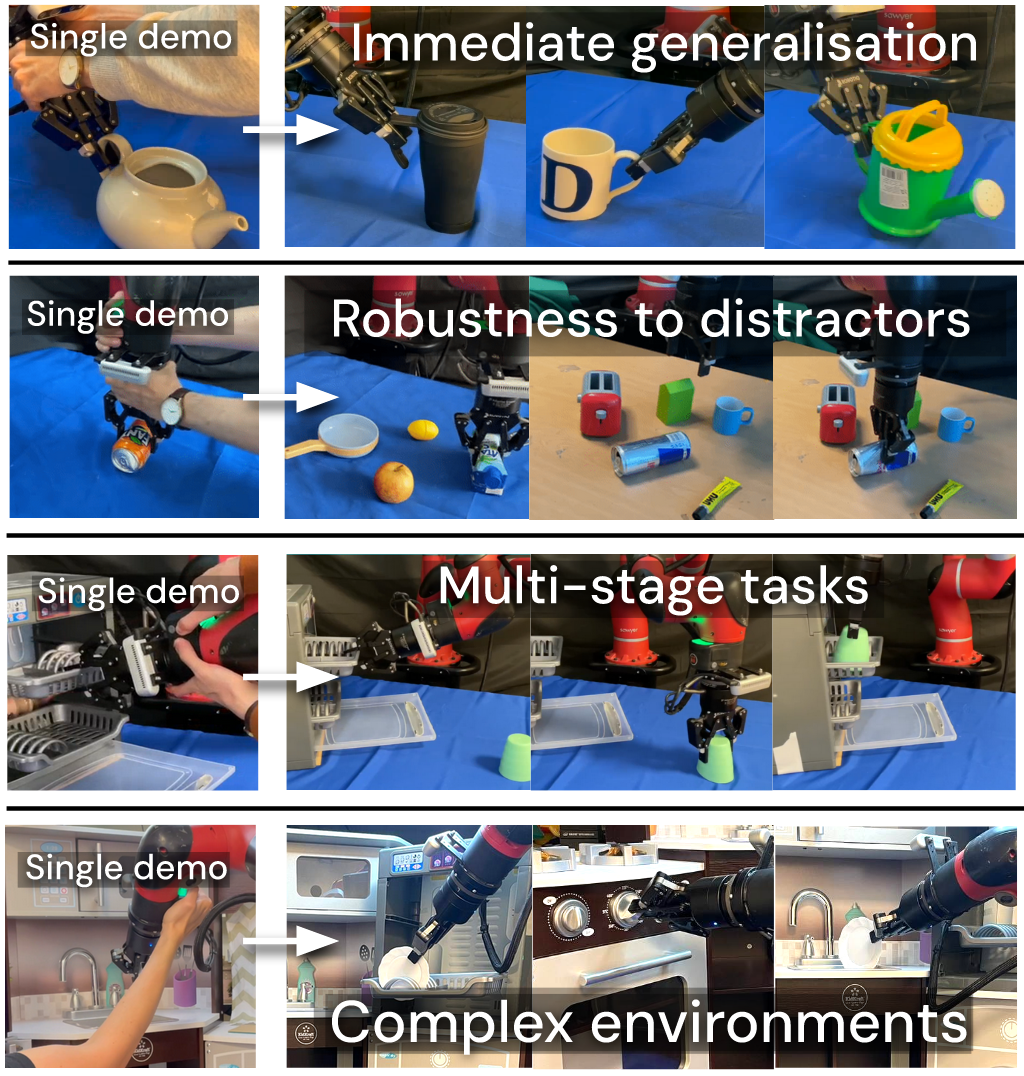}
    \end{center}
    \caption{From a \textbf{single demo}, DINOBot can learn to adapt to new objects, be robust to distractors, execute multi-stage long horizon tasks, and interact with complex environments.}
    \label{fig:obj_generalisation}
\end{figure}

\begin{figure*}[t!]
    \centering
    \includegraphics[width=.9\textwidth]{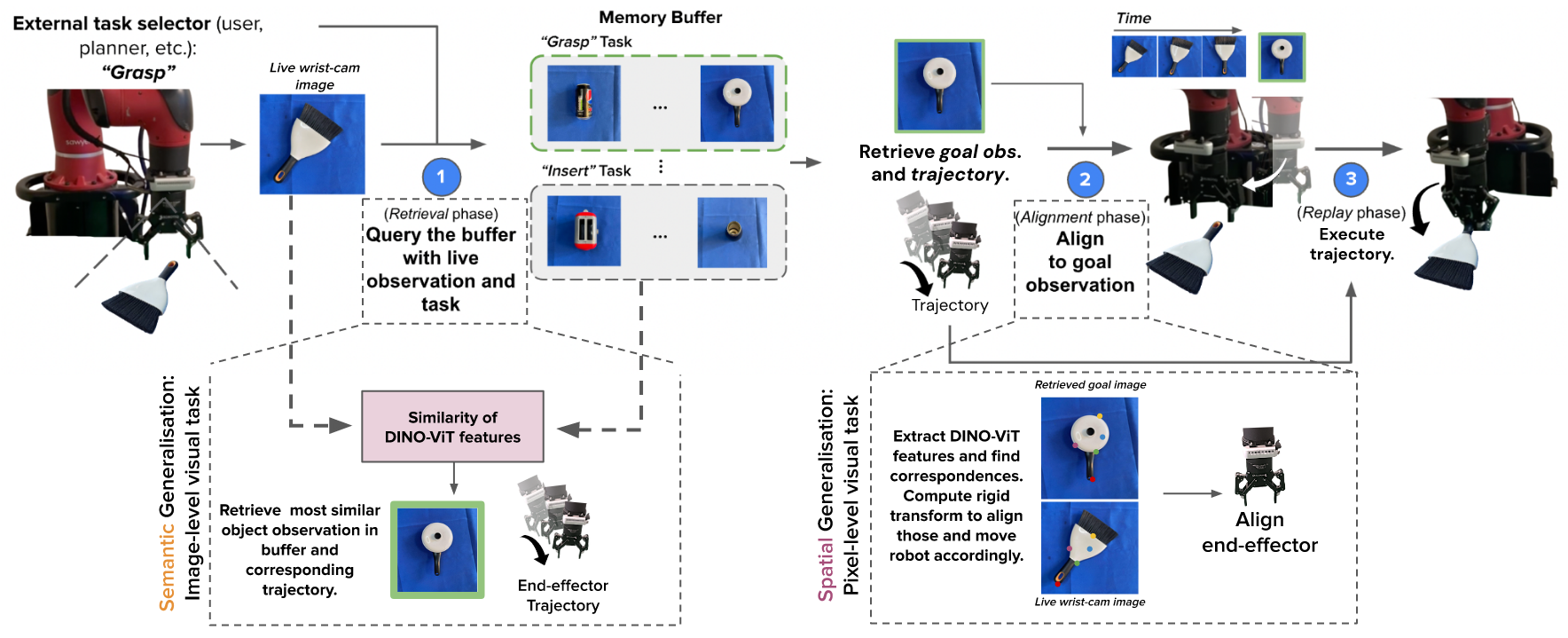}
    \caption{Overall illustration of our framework. Upon observing a new object, the robot visually compares it other objects observed during demonstrations to find the most similar object (\textit{semantic, image-level reasoning}), and retrieve both its image and the trajectory executed on that object. Then, the robot aligns its end-effector with this image (\textit{spatial, pixel-level reasoning}), before then executing that trajectory. These two phases of reasoning are both based on extracting and matching DINO-ViT features.}
    \label{fig:key-idea}
     \label{fig:retrieval}
     \vspace{-10pt}
\end{figure*}

As described by DINO's authors, DINO-ViTs extract ``\textit{universal features suitable for image-level visual tasks as well as pixel-level visual tasks}" \cite{oquab2023dinov2}. Inspired by these two capabilities, we designed DINOBot around two distinct modes of reasoning. Firstly, image-level \textit{semantic} reasoning to generalise learned behaviours to novel objects. Secondly, pixel-level \textit{geometric} reasoning to generalise learned behaviours to novel object poses. We integrate these two modes by modelling a manipulation task as a semantic image \textit{retrieval} task, followed by a geometric \textit{alignment} task (Fig. \ref{fig:key-idea}). 

Through a series of real-world experiments, studying a range of tasks such as grasping, pouring, and inserting objects, we show not only that DINOBot achieves one-shot imitation learning on tasks where existing methods require many demonstrations, but it also achieves very efficient generalisation to novel objects. Our conclusion, and the key takeaway message, is clear: \textbf{designing an imitation learning framework around image-level (retrieval) and pixel-level (alignment) visual tasks allows us to leverage the remarkable capabilities of DINO-ViTs, leading to unprecedented learning efficiency compared to alternative paradigms}.   



\section{Related Work}
\label{sec:related}

\textbf{Imitation Learning on top of Foundation Models.} Many recent imitation learning methods for robot manipulation train end-to-end policies, built upon features extracted from pre-trained models. However, such methods still require tens or hundreds of demonstrations per task \cite{shah2021rrl, nair2022r3m, radosavovic2023real}.  In our work, we show that an explicit decomposition into an image-level retrieval phase and a pixel-level alignment phase followed by a trajectory replay,  results in a substantial improvement in data and time efficiency.

\textbf{Retrieval for few-shot learning}. Related to our use of DINO's image-level capabilities, retrieval for few-shot learning, especially in robotics, reinforcement learning and control, has also been investigated elsewhere \cite{lu2020learning, boney2017semi, 1570827, Sharon2005SynthesisOC, shah2018q, mansimov2018simple, pari2021surprising, du2023behavior, retrieval_flowcontrol}. Some approaches directly retrieve the actions to execute \cite{mansimov2018simple, pari2021surprising, Sharon2005SynthesisOC} or the data to train a policy \cite{du2023behavior}. \cite{retrieval_flowcontrol} also retrieves the most similar demonstration from a buffer, and then aligns the robot by using optical flow between the current and goal observations, but is considerably worse at generalising to new objects as we demonstrate in Sec. \ref{sec:experiments}. 

\textbf{Local correspondences for robot manipulation}. Related to our use of DINO's pixel-level capabilities, extraction of dense or sparse correspondences through neural networks has been explored elsewhere for robot manipulation \cite{florence2018dense, manuelli2022kpam, vecerik2021s3k}. However, these methods require object-specific data collection to train networks. \cite{vecerik2023robotap} uses a general keypoint-tracking network, that however does not generalise to unseen objects as we will illustrate in later sections. \cite{goodwin2022you, hadjivelichkov2023one, amir2021deep} demonstrate the capabilities of off-the-shelf DINO-ViT features for pose estimation. On top of these, we build a full IL framework based on retrieval and alignment. \cite{vosylius2023few} trains a graph neural networks that takes as input local geometry embeddings of point clouds to perform few-shot IL. \cite{wang2023d3fields,ju2024roboabc,wang2023sparsedff} also use DINO features to find matches and keypoints, they however mostly focus on grasping, while DINOBot can learn a large repertoire of everyday skills.

\textbf{Trajectory decomposition for robot manipulation}. Our work is related to \cite{johns2021coarse, valassakis2022demonstrate, vitiello2023one}, which also decompose robot manipulation into visual alignment and then replay of a trajectory. \cite{johns2021coarse} trains an object-specific visual servoing policy with autonomously collected data, while \cite{valassakis2022demonstrate} trains a general goal-conditioned alignment policy in simulation. However, neither of these methods enable generalisation to novel objects, and furthermore, with DINOBot we demonstrate that, by using DINO-ViTs, neither object-specific data collection nor additional simulation training is needed. 

\textbf{Semantic and spatial reasoning}. CLIPort \cite{shridhar2022cliport}, like DINOBot, also phrases object manipulation as semantic reasoning combined with spatial reasoning. However, there are some fundamental differences: 1) their pipeline is more implicit, going from language and visual observations to affordance prediction through a single forward pass of a two-streams network, while we have explicit retrieval and alignment phases, that leads to DINOBot's better efficiency 2) their method is designed for top-down pick-and-place-like tasks, while we experiment with more complex tasks, also in a 6-DOF environment like a kitchen.

\section{Method}
\label{sec:method}
\label{sec:preliminaries}

Figure \ref{fig:key-idea} illustrates our framework. \revision{During deployment, DINOBot executes manipulation tasks} via a semantic \textit{retrieval} and a spatial \textit{alignment }phase, followed by a demonstration \textit{replay} phase. During training, the operator provides demonstrations, and the collected data fills a memory buffer. For each demonstration, this data consists of a wrist-camera image captured at the beginning of the demonstration, and the end-effector trajectories for the demonstration. During testing, for the \textit{retrieval} phase, the robot queries the memory buffer with its current wrist-camera observation. Given the nature of the task to be executed (e.g. ``Grasp" or ``Insert"), it retrieves the most similar image which is then used as goal image (Fig. \ref{fig:correspondences}), and the corresponding end-effector trajectory. The interaction with the object is decomposed into the \textit{alignment} and \textit{replay} phases. It first uses the retrieved goal image to align the end-effector with the test object correctly. Then the robot replays the recorded demonstration trajectory. Our method only needs an RGB-D wrist camera, rigidly mounted to the robot's end-effector. No prior knowledge of objects or tasks is needed, and no external camera is needed.

\subsection{\textbf{Spatial Generalisation through Alignment and Replay}}
\label{sec:how}
\label{sec:where}

To teach the robot how to interact with an object, the human operator manoeuvres the end-effector to provide a demonstration, e.g. with kinesthetic teaching. The end-effector starts from the \textit{bottleneck pose}, $B^O$, \revision{which is a pose arbitrarily chosen by the operator to start the demonstration from, and from where the object must be visible from the wrist camera}. The demonstration is recorded as a sequence of 3D linear and 3D angular velocities of the end-effector, expressed in the end-effector frame $E$. This sequence is the trajectory $s$, such that $ s = [\mathcal{V}^E_1, \dots, \mathcal{V}^E_{T}]$, where $\mathcal{V} = [v_x, v_y, v_z, w_x, w_y, w_z]$. When starting the demonstration from pose $B^O$, the robot also records the wrist-camera observation of the object from that pose, which we call the \textit{bottleneck observation}. For each demonstration, a \textit{task} name is also specified by the operator, e.g. ``Grasp", ``Insert", or ``Pour". In summary, each time a new demonstration is recorded by the operator, the framework adds the following data to the memory buffer: the bottleneck observation recorded before starting the demonstration, the trajectory of velocities $ s = [\mathcal{V}^E_1, \dots, \mathcal{V}^E_{T}]$, and the task name.

We can imagine the bottleneck pose being rigidly attached to the virtual (since we do not assume object models) object frame. With $W$ being the world frame, when we move the object, the global bottleneck pose $B^W$ moves rigidly with the object, whilst the local bottleneck pose $B^O$ stays constant. If the end-effector is re-aligned to $B^O$ when the object is moved, replaying the end-effector velocities of the recorded trajectory would suffice in solving the task \cite{johns2021coarse}. Thus, the robot only needs to reach $B^O$ again for novel poses of this object, which is now expressed as a different $B^W$.

\begin{wrapfigure}{l}{0.3\textwidth}
    \begin{center}
    \includegraphics[width=0.3\textwidth]{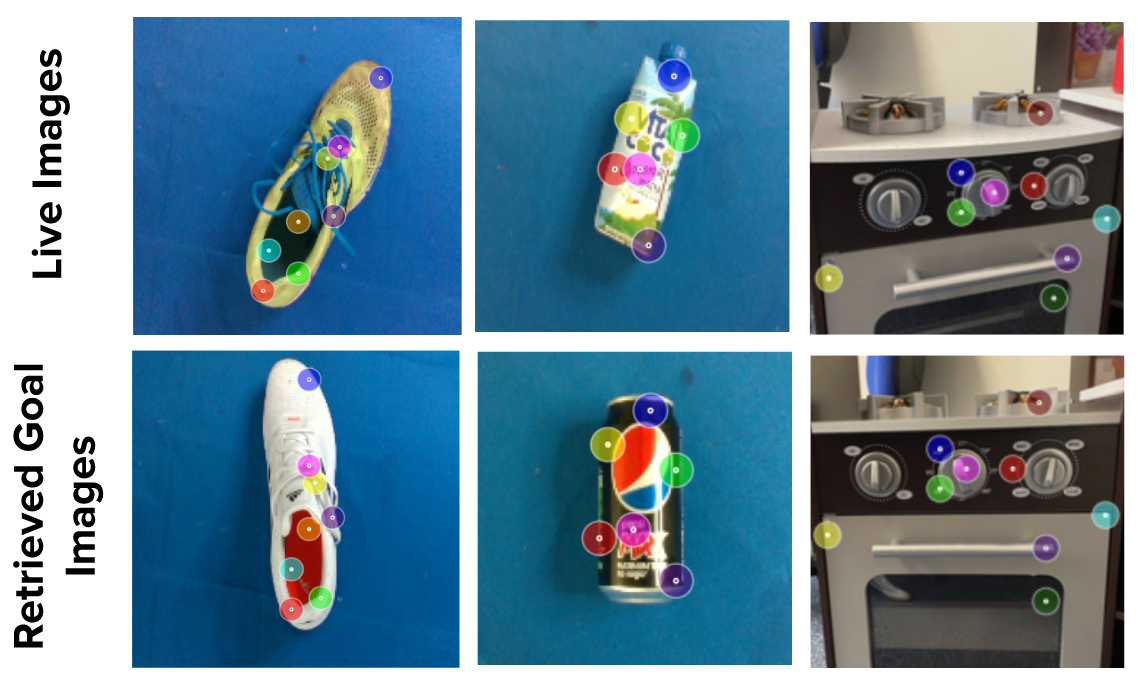}
    \end{center}
    \caption{In each column, given a live image (top) DINOBot retrieves from the buffer the most similar image in the buffer (bottom), and finds correspondences between the two.}
    \label{fig:correspondences}
\end{wrapfigure}

When a new object is observed during testing, how can the robot align itself with the object to reach $B^W$ again? Given the observation recorded at the beginning of some demonstration stored in the buffer, and the current live observation recorded from the wrist-camera, DINOBot performs visual servoing to align the two images. Specifically, we extract deep patch features \cite{dosovitskiy2021an} from the DINO-ViT and use the method described in \cite{amir2021deep} to find correspondences through a Best Buddies Nearest Neighbour matching phase \ref{fig:correspondences}). This procedure generates two lists of keypoints, defined as 3D coordinates $C_1 = \{x_{1,i}, y_{1,i}, z_{1,i}\} $, $C_2 = \{x_{2,i}, y_{2,i}, z_{2,i} \}$ (we extract the depth of each RGB pixel correspondence through the RGB-D wrist camera). We then compute the least-squares rigid transformation that aligns the two lists of corresponding keypoints \cite{sorkine2017least, svd} and move the robot accordingly. We repeat this process until the alignment is precise enough, i.e. the norm of the distance of the two lists of correspondences is smaller than a threshold. Once the alignment is completed, replaying the recorded trajectory allows the robot to successfully interact again with the object. We provide additional info and code on the website.

\subsection{\textbf{Semantic Generalisation through Retrieval}}
\label{sec:what}

In the previous section, we described how we formulate the interaction with an object as visual alignment with a goal image, followed by replay of a trajectory. But given a new object to interact with, how does the robot select the best goal image and trajectory from its memory buffer? First, we assume that the robot receives, either by the human operator or by an external planner, the \textit{task} to execute with the object, e.g. ``Grasp" or ``Open". The robot then records a live observation of the object from its wrist-camera. It then \textit{retrieves} from its memory buffer the \textit{bottleneck observation} most similar to the live observation, from the subset of the memory buffer for that task (e.g. if the task is to ``Grasp", then all demonstrations which were also ``Grasp" tasks would be considered). DINOBot performs retrieval by extracting features for each observation in the buffer and the live observation through the DINO-ViT. In our case, we extract the CLS token, a $1 \times 768$ vector \cite{dosovitskiy2021an}. It then performs a nearest neighbour search by computing the cosine similarity between the extracted features, as described in \cite{oquab2023dinov2}, to find the closest match in the buffer. The best match is retrieved together with its recorded trajectory: these are then used as goal observation for the \textit{alignment} phase and actions to be replayed during the \textit{replay} phase. As such, DINOBot is able to generalise observed demonstrations to novel objects, by independently using both the image-level and pixel-level capabilities of vision foundation models. 


\begin{figure*}[t!]
    \begin{center}
    \includegraphics[width=.95\textwidth]{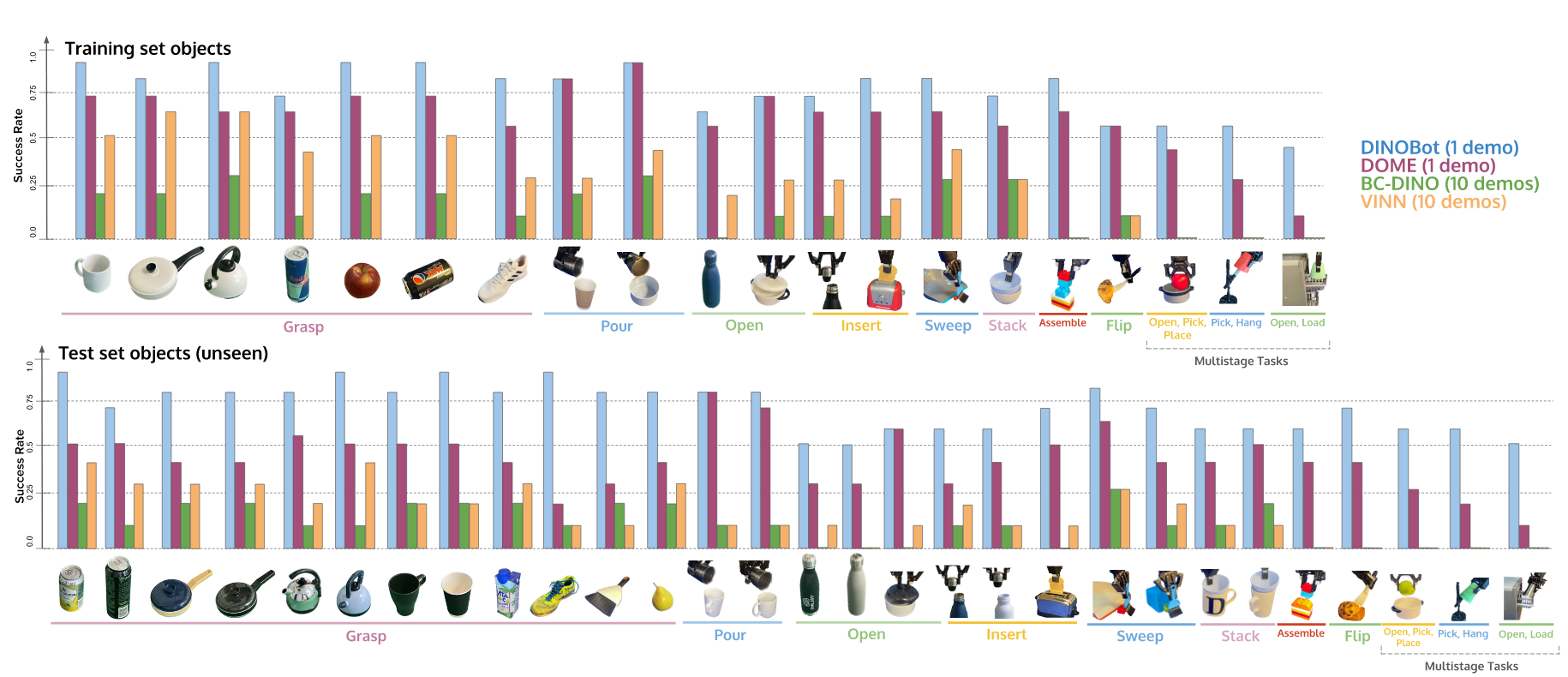}
    \end{center}
    \caption{\revision{Success rates on each object for all methods.}}
    \label{fig:full_results}
    \label{fig:all_testing}
\end{figure*}

\section{Experiments}
\label{sec:experiments}
\label{sec:baselines}

In this section, we empirically measure the ability of our method to learn behaviours efficiently and transfer them effectively to new objects, and we structure our experiments around several important questions which we present in the following pages. \revision{These experiments were conducted on two different real-world environments, \revision{on a total of \textbf{15 tasks} with \textbf{53 objects}} (a task, in this work, can be intuitively described as a verb, like \textit{grasp} or \textit{open}, and be applied to many different objects). First, a tabletop environment with 49 everyday objects (20 train, 29 test) and 8 types of tasks: \textbf{grasping}, \textbf{pouring}, \textbf{opening}, \textbf{inserting}, \textbf{sweeping}, \textbf{stacking}, \textbf{assembling}, \textbf{flipping} and 3 \textit{multi-stage tasks} following the demonstration procedure in \cite{di2022learning}: \textbf{open+pick+place}, \textbf{hanging cups} and \textbf{opening and loading a dishwasher}. Second, a toy kitchen environment where we teach 4 tasks: \textbf{opening} the microwave, \textbf{inserting} a plate into a dishwasher, \textbf{placing} a plate in the sink, and \textbf{turning} an oven knob.} 
To prove the effectiveness and generality of DINOBot, we took these tasks from other recent imitation learning papers, and we therefore test on \textit{all the tasks} from: 1) \textbf{Relational-NDF} \cite{simeonov2023se}, (where instead of \textit{bottle in container} we have multiple, more precise insertion tasks) + 2) \textbf{FISH} \cite{haldar2023teach}, (non multifingered ones - as \textit{door opening}, we have \textit{open a microwave door} - for \textit{key insertion}, we have several precise insertion tasks with a lower than 5mm error tolerance), + 3) \textbf{VINN} \cite{pari2021surprising}, (where instead of \textit{pushing} directly, we sweep, i.e. push with a tool), + 4) \textbf{Relay Policy Learning} \cite{gupta2019relay}. 
We invite you to watch the videos on our website at \href{https://sites.google.com/view/dinobot}{https://sites.google.com/view/dinobot}.\\

\textbf{Setup}: We run our experiments on a Sawyer robot mounting a Robotiq 2F-85 gripper. We use a single wrist-mounted Intel RealSense D435 camera. The camera receives RGB-D images which we rescale to $224\times224\times4$. The robot starts learning \textit{tabula rasa}: no previous knowledge of tasks or objects, such as CAD models, is used. 

\textbf{Baselines}: We compare DINOBot to the following baselines. 1) \textbf{DOME} \cite{valassakis2022demonstrate}, a similar one-shot IL method based on a learned (through simulation), goal-conditioned visual servoing phase followed by a trajectory replay. However, DOME does not generalise to novel objects, and so we extended the original method with our retrieval phase to provide the goal image to align with. This baseline exists to compare DINO's pixel-level alignment abilities with DOME's simulation-trained visual servoing network. 2) \textbf{BC-DINO}, a Behaviour Cloning (BC) implementation that trains a network to output actions on top of features extracted through a DINO-ViT. This baseline exists to compare our framework to the more typical approach in recent works, where BC policies are trained upon features extracted from pre-trained models \cite{nair2022r3m, shah2021rrl, radosavovic2023real} on all the collected demonstrations. \revision{3) \textbf{VINN}: Visual Imitation through Nearest Neighbours (VINN) \cite{pari2021surprising}, frames Learning from Demonstration entirely as a retrieval problem, where actions are computed by retrieving the $k$ most similar demo observations and averaging their corresponding actions. We use DINO features to embed the observations and retrieve them.}

All methods receive as input the task to execute and the wrist-camera observation. For DINOBot and DOME, we provide a single demonstration for each training object. As BC-DINO and VINN are not designed as one-shot IL methods, we provide 10 demonstrations per training object. More details on the implementation of these methods and on the experiments can be found on our website.


\label{sec:general_exp}

\textbf{\\Can DINOBot learn everyday-like tasks efficiently, and transfer those skills to novel objects? How does it compare to baselines from recent literature?}
To answer this, we train each method with demonstrations of how to interact with the objects in the training set on the tabletop environment (Fig. \ref{fig:all_testing}, top). We then test each method using the objects depicted in the figure: both the training objects (top), and the unseen test objects (bottom). At test time, we position each object randomly on the table, and inform the robot what the task to perform is (e.g. ``Grasp", ``Open", etc). We sample a position inside a 40cm $\times$ 40cm area, and an angle between -45° and 45° relative to the original demonstration angle. We run 10 trials per object, sampling a new object pose each time. In this tabletop scenario, the alignment phase is 4-DOF, with the robot always aligned with the vertical axis and only rotating around this axis, while the \textit{replay phase} trajectory execution is 6-DOF. Later in the paper, we will also explore 6-DOF alignment in the kitchen environment.

The performance of each method on each object is shown in Fig. \ref{fig:all_testing}. We group the results into training set objects (for which the robot received demonstrations, to study one-shot IL), and test set objects (to study generalisation to novel objects).

The results show that not only can DINOBot obtain remarkable one-shot performance on the training objects, but that it can generalise to unseen objects: performance stays considerably close to the training set performance.
The performance of the baselines is noticeably lower, illustrating the benefits of our combination of retrieval and alignment to leverage the capabilities of DINO-ViTs. Using the extracted features as input to train a BC network, perhaps the most typical paradigm for using pre-trained networks in IL \cite{nair2022r3m, radosavovic2023real, shah2021rrl}, clearly performs worse, even with 10 times the number of demonstrations. DOME performs well on training set objects, but performance degrades when trying to generalise to unseen objects, as the method was not designed to do so. 

\label{sec:kitchen}
\textbf{\\Can DINOBot learn to interact with a complex kitchen environment?}  In this section, we now study more complex 6-DOF tasks and environments. We provide a single demonstration to our robot for the following tasks: 1) open a microwave, 2) insert a plate into the dishwasher, 3) put a plate into the sink, and 4) turn an oven knob. After each demonstration, we test the ability of our method to replicate the task with the robot starting from a different starting position, with 10 trials for each task. As the only input the robot receives are wrist-camera observations and no proprioceptive data, moving the robot to a different initial state is akin to moving the kitchen to test for spatial generalisation. In this experiment, we skip the \textit{retrieval} phase to study the \textit{alignment} and \textit{replay} phases' one-shot IL ability in isolation, such that the robot is asked to perform the same task it has just been shown from the demonstration. The robot now has to perform a challenging 6-DOF alignment by extracting correspondences between the \textit{bottleneck observation}, recorded at the start of the demonstration, and its live wrist-camera observation (videos of execution and keypoints extractions on our website). We compare DINOBot with one demonstration to \textbf{BC-DINO} \revision{and \textbf{VINN}} with 10 demonstrations. DOME is designed for 4-DOF tabletop settings and cannot be deployed for this task.

\begin{wrapfigure}{l}{0.3\textwidth}
    \begin{center}
    \includegraphics[width=0.3\textwidth]{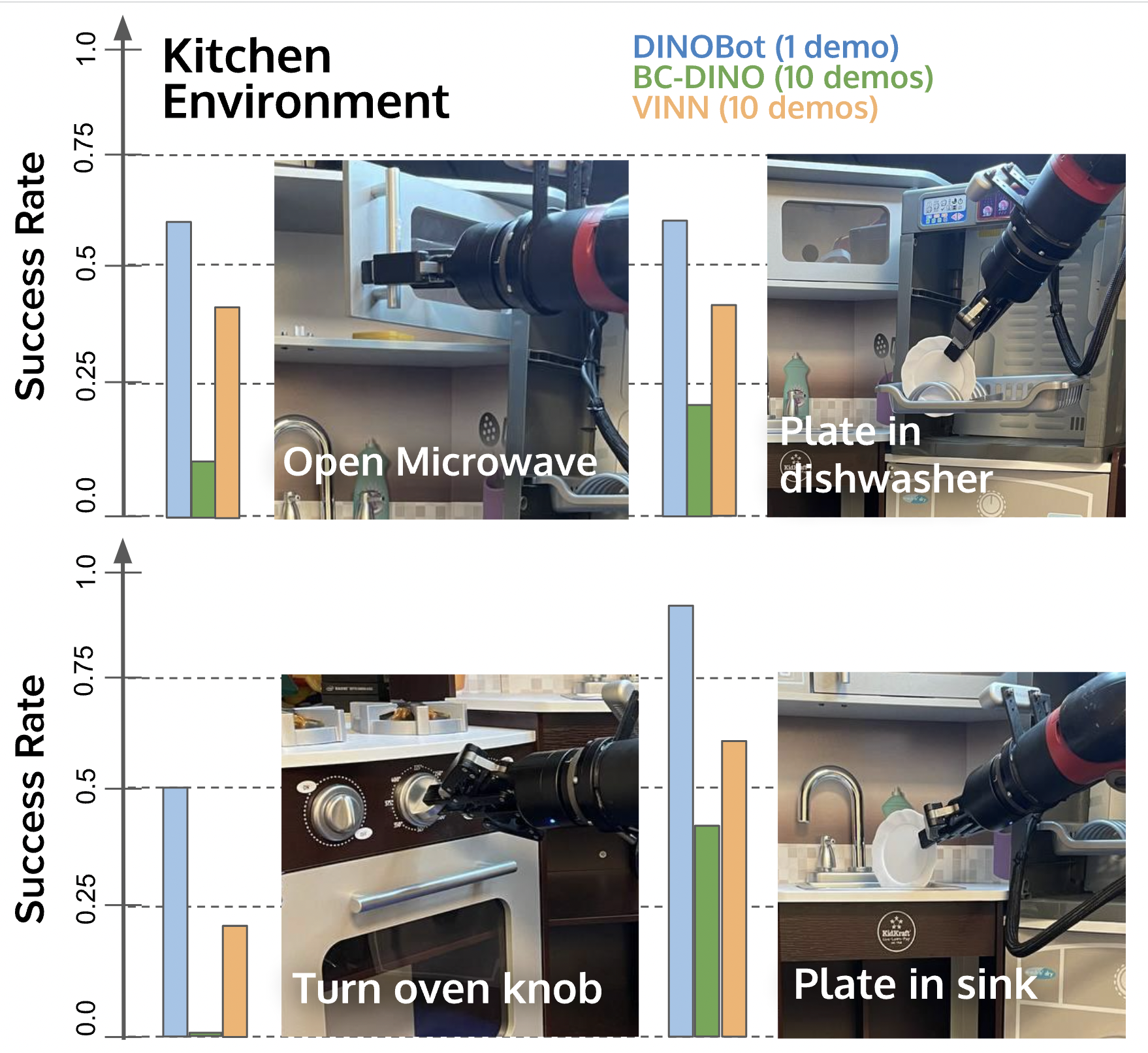}
    \end{center}
    \caption{Results on kitchen tasks.}
    \label{fig:kitchen_hor}
\end{wrapfigure}

Results shown in Fig. \ref{fig:kitchen_hor} clearly show that our framework can successfully solve these everyday-like tasks, when receiving only a single demonstration. As with the previous experiments, the significantly better performance than BC-DINO and VINN again shows that explicitly using DINO's extracted visual features is superior to a more standard use of DINO as just backbone features.

\begin{table}[b]
    \centering 
    \begin{tabular}{ccc}
        \hline
         Method&  w/out Distractors& w Distractors\\ \hline 
         \textbf{DINOBot (1 demo)}&  \textbf{0.8}& \textbf{0.76}\\ 
         \textbf{DOME (1 demo)}&  0.67& 0.5\\ 
         \textbf{BC-DINO (10 demos)}&  0.2& 0.07\\ 
         \textbf{VINN (10 demos)}&  0.33& 0.07\\ 
    \end{tabular}
    \caption{Success rates with and without distractors.}
    \label{table:distractors}
\end{table}

\begin{wrapfigure}{r}{0.27\textwidth}
    \begin{center}
    \includegraphics[width=0.27\textwidth]{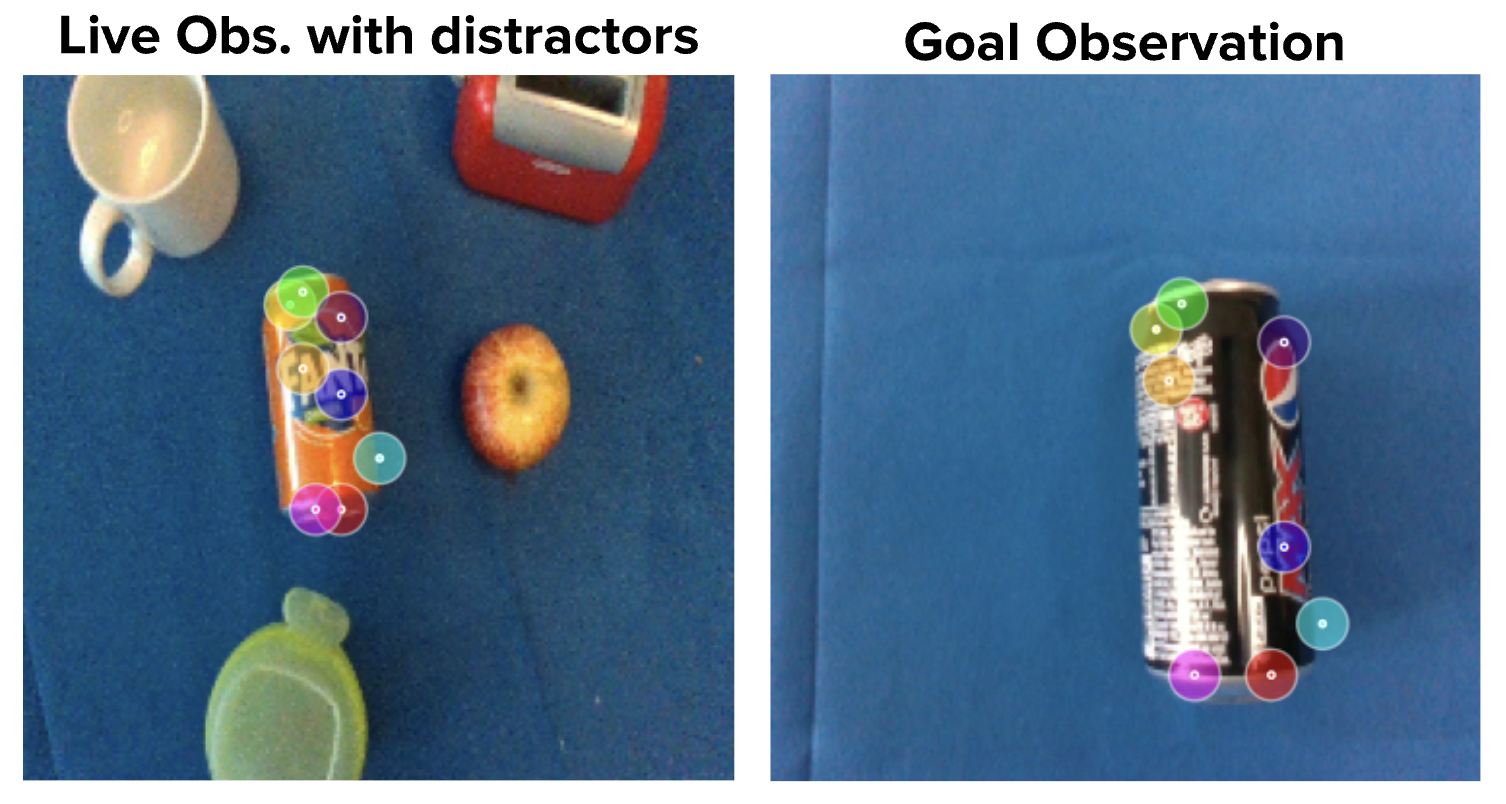}
    \end{center}
    \caption{Despite the distractors, sensible correspondences can still be found.}
    \label{fig:distractors}
\end{wrapfigure}

\textbf{\\Does DINOBot work even in the presence of distractor objects?} In this experiment, we provide a single demonstration for each of the following tasks: 1) \textbf{grasping} a can, 2) \textbf{inserting} bread into a toaster, and 3) \textbf{pouring} from a cup into a mug.  Here we study the \textit{alignment} and \textit{replay} phases robustness to distractors in isolation, hence we provide the goal image manually, skipping the \textit{retrieval} phase. At test time, we use an unseen test set object (e.g. a different can or a different toaster), but we also vary the scene from the demonstration setup by adding a set of distractors from different classes (Fig. \ref{fig:distractors}). We compare DINOBot's performance against the baselines: DOME receives a single demonstration as well, while BC-DINO and VINN receive 10 demonstrations.

We run 10 test trials for each task, with and without distractors. Results in Table \ref{table:distractors} demonstrate that DINOBot not only surpasses all the baselines, but faces the smallest decrease in performance due to distractors. This demonstrates that DINO-ViT is very robust and able to extract correct correspondences between the goal observation object and the live observation, even in the presence of additional distractor objects(Fig. \ref{fig:distractors}).

\label{sec:compare_retrieval}
\textbf{\\What technique performs best in retrieving the most similar observation from the memory buffer?} Being an essential part of our method (Sec. \ref{sec:what}), we investigate the performance of different retrieval techniques to further motivate our design choice. 

\begin{wrapfigure}{r}{0.25\textwidth}
    \begin{center}
    \includegraphics[width=0.25\textwidth]{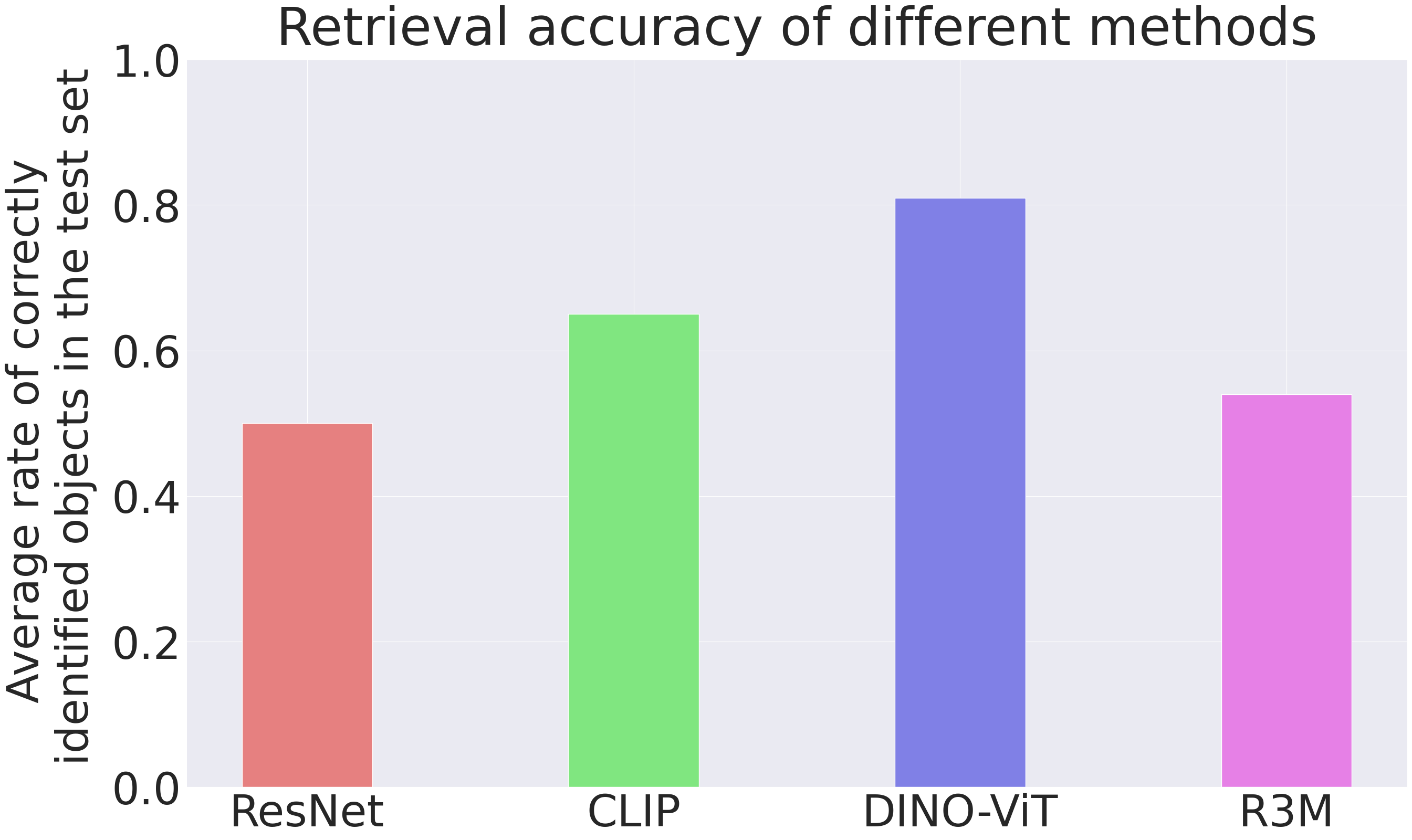}
    \end{center}
    \caption{Comparison of retrieval accuracy of the baselines.}
    \label{fig:retr_accuracy}
\end{wrapfigure}

We compare retrieval based on features extracted from publicly-available ImageNet-trained ResNet-50 \cite{he2016deep}, CLIP \cite{radford2021learning}, R3M \cite{nair2022r3m} and DINO-ViT \cite{caron2021emerging}, each of which has been pre-trained on vast datasets of images (refer to the website for additional details). We measure retrieval accuracy by providing to each method an observation of an unseen object from the test set, composed of the 20 objects of Fig. \ref{fig:all_testing} (bottom). We then measure how many times the method retrieves from the buffer an observation of the object belonging to the same class. We show results in Figure \ref{fig:retr_accuracy}. The features extracted via a DINO-ViT achieve the best accuracy, confirming recent findings from the literature \cite{caron2021emerging, amir2021deep} that such features effectively encode semantic, geometric, global and local information of the observation.

\textbf{How accurate is the keypoints-based alignment phase of DINOBot?} 
Here, we evaluate alignment precision in isolation, comparing it to two other alignment techniques from recent literature: \textbf{RoboTAP} \cite{vecerik2023robotap}, which uses keypoints extracted through \textbf{TAPIR} \cite{doersch2023tapir} for visual servoing and alignment, and \textbf{FlowControl} \cite{argus2020flowcontrol, retrieval_flowcontrol}, which uses optical flow to perform alignment; for the optical flow network, we use \textbf{RAFT} \cite{teed2020raft}. 
We use three pairs of objects: shoes, cans, and toasters. We generate a series of top-down observations of the objects where we manually translate and rotate the objects, in addition to artificially changing the background. Each method is then tested on its accuracy to compute the correct translation and rotation given an initial observation and a transformed observation. We test each method both when the pair of images depict the same object, and when the two images contain two different objects of the same class, e.g. two different shoes, to evaluate for semantic and geometric robustness. 

\begin{wrapfigure}{l}{0.33\textwidth}
    \begin{center}
    \includegraphics[width=0.33\textwidth]{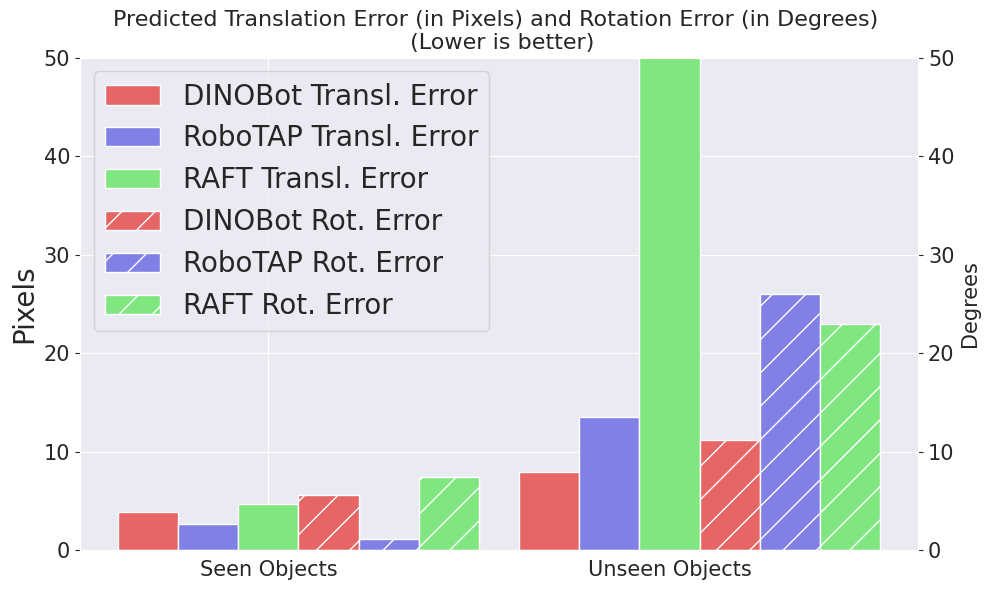}
    \end{center}
    \caption{Alignment accuracy benchmark.}
    \label{fig:alignment_bench}
\end{wrapfigure}

Results in Fig. \ref{fig:alignment_bench} demonstrate that, while RoboTAP is extremely accurate when the same object appears in the pair of observations, performance degrades considerably when matching two different objects, albeit belonging to the same class. RAFT shows a similar trend, although with overall worse performance. 
DINOBot instead is able to generalise to unseen objects, maintaining strong performance.

\textbf{\\How much can DINOBot generalise to objects with different sizes and appearance from a single demo?} Here, we collected a new set of objects belonging to 4 classes - bottles, kettles, cups, and pans - where the object sizes and shapes vary significantly within each class. For each class, we then provide a demo for a single object, and then test on all the objects - 1 demo object, and 4 unseen objects - measuring the success rate over 10 test trials for each. As such, here we investigate the ability to generalise to novel objects that can vary substantially in size and shape, e.g. bottles more than double in size, mugs with handles of different shapes and position, etc. Results in Figure \ref{fig:obj_generalisation} show that DINOBot achieves significant generalisation and again outperforms baselines. 

\begin{figure}[t]
    \begin{center}
    \includegraphics[width=0.4\textwidth]{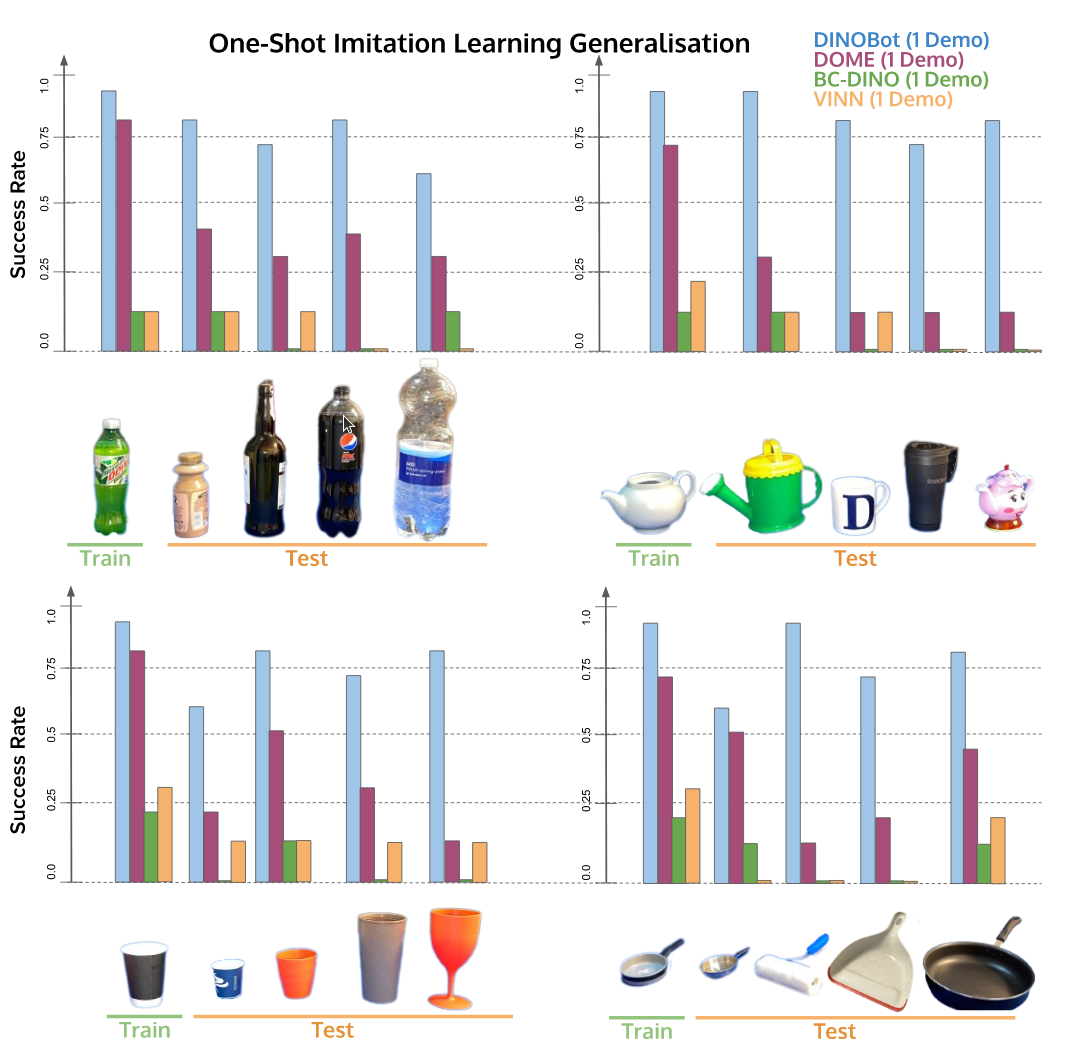}
    \end{center}
    \caption{One-shot, intra-class IL generalisation experiments.}
    \label{fig:obj_generalisation}
    \vspace{-5pt}
\end{figure}

\textbf{\\How does DINOBot compare to other methods across three challenges: adaptability, dexterity, and precision?} To better highlight the different manipulation abilities expressed by our method, we cluster the results for the tasks in \ref{fig:all_testing} into three groups, which cover some of the main abilities a robot should possess to tackle everyday-like tasks: \textbf{adaptability}, \textbf{dexterity}, and \textbf{precision}. 

\begin{wrapfigure}{r}{0.3\textwidth}
    \begin{center}
    \includegraphics[width=0.3\textwidth]{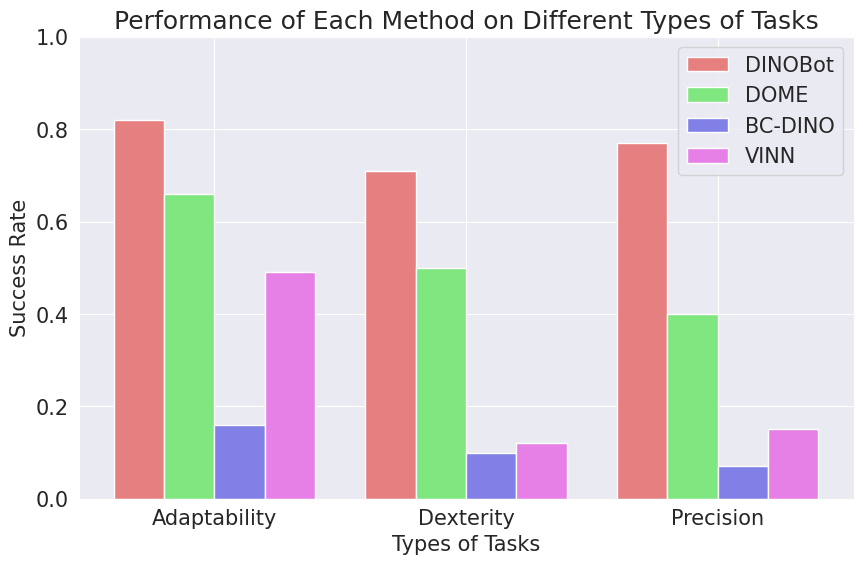}
    \end{center}
    \caption{Performance of each method on tasks that require \textbf{adaptability}, \textbf{dexterity} or \textbf{precision.}}
    \label{fig:task_groups}
\end{wrapfigure}

We cluster as follows. 1) All results from \textit{grasp} experiments are grouped into \textbf{adaptability}, as grasping is the task that encounters the largest variety in objects \cite{Fang2020GraspNet1BillionAL}, and thus requires significant adaptation to novel shapes based on visual observation. 2) All results from \textit{pour, sweep, flip, open dishwasher} are grouped into \textbf{dexterity}, as these require non-trivial, often fast movements to succeed. 3) All results from \textit{insert, assemble} are grouped into \textbf{precision}, as these have lower than 5mm positional tolerance. Results, illustrated in Fig. \ref{fig:task_groups}, demonstrate that DINOBot surpasses all baselines on all these challenges. We observed that the use of retrieval helped DINOBot adapt to unseen objects of various shape, while alignment and trajectory replay helped tackling precise and dexterous tasks, strongly surpassing the end-to-end baselines.

\section{Conclusion}
\label{sec:final_discussion}

We have introduced DINOBot, an imitation learning framework designed around image-level (retrieval) and pixel-level (alignment) visual tasks, to take full advantage of the abilities of DINO-ViT foundation models. Our extensive experimental investigation demonstrated how this framework surpasses other DINO-based baselines, that do not harness the full capabilities of these vision models. On our website we list a series of limitations and additional questions and visualisations that may provide additional insights to the reader.

\newpage

\bibliographystyle{IEEEtran}
\bibliography{paper}

\end{document}